\documentclass{article}

\usepackage[final]{corl_2020}
\usepackage{todonotes}
\usepackage{amsmath,amsfonts,amssymb}
\usepackage[ruled,vlined]{algorithm2e}
\usepackage{graphicx}
\usepackage{mdframed}
\usepackage{subcaption}
\usepackage{wrapfig}
\usepackage{multirow}
\usepackage{xcolor}
\usepackage{color, colortbl}
\usepackage[noadjust]{cite}
\usepackage{hyperref}
\usepackage{siunitx}
\usepackage{stackengine,scalerel}

\definecolor{Gray}{gray}{0.9}
\definecolor{LightCyan}{rgb}{0.88,1,1}

\newcommand{\mb}[1]{\mathbf{#1}}
\newcommand{\tb}[1]{\textbf{#1}}
\newcommand{\bs}[1]{\boldsymbol{#1}}

\newcommand\overstar[1]{\ThisStyle{\ensurestackMath{%
  \setbox0=\hbox{$\SavedStyle#1$}%
  \stackengine{0pt}{\copy0}{\kern.2\ht0\smash{\SavedStyle*}}{O}{c}{F}{T}{S}}}}

\title{Learning Predictive Models for Ergonomic Control of Prosthetic Devices}

\author{
  Geoffrey Clark\\
  Interactive Robotics Lab\\
  Arizona State University\\
  \texttt{gmclark1@asu.edu} \\
  \And
  Joseph Campbell\\
  Interactive Robotics Lab\\
  Arizona State University\\
  \texttt{jacampb1@asu.edu} \\
  \And
  Heni Ben Amor\\
  Interactive Robotics Lab\\
  Arizona State University\\
  \texttt{hbenamor@asu.edu} \\
}

\begin{document}
\maketitle

\begin{abstract}
We present Model-Predictive Interaction Primitives -- a robot learning framework for assistive motion in human-machine collaboration tasks which explicitly accounts for biomechanical impact on the human musculoskeletal system. 
First, we extend Interaction Primitives to enable \emph{predictive biomechanics}: the prediction of future biomechanical states of a human partner conditioned on current observations and intended robot control signals. 
In turn, we leverage this capability within a model-predictive control strategy to identify the future ergonomic and biomechanical ramifications of potential robot actions. Optimal control trajectories are selected so as to minimize future physical impact on the human musculoskeletal system. 
We empirically demonstrate that our approach minimizes knee or muscle forces via generated control actions selected according to biomechanical cost functions. Experiments are performed in synthetic and real-world experiments involving powered prosthetic devices.
\end{abstract}

\section{Introduction}
\begin{wrapfigure}{r}{0.4\textwidth} 
\centering
\vspace*{-\baselineskip}
\includegraphics[width=0.4\textwidth]{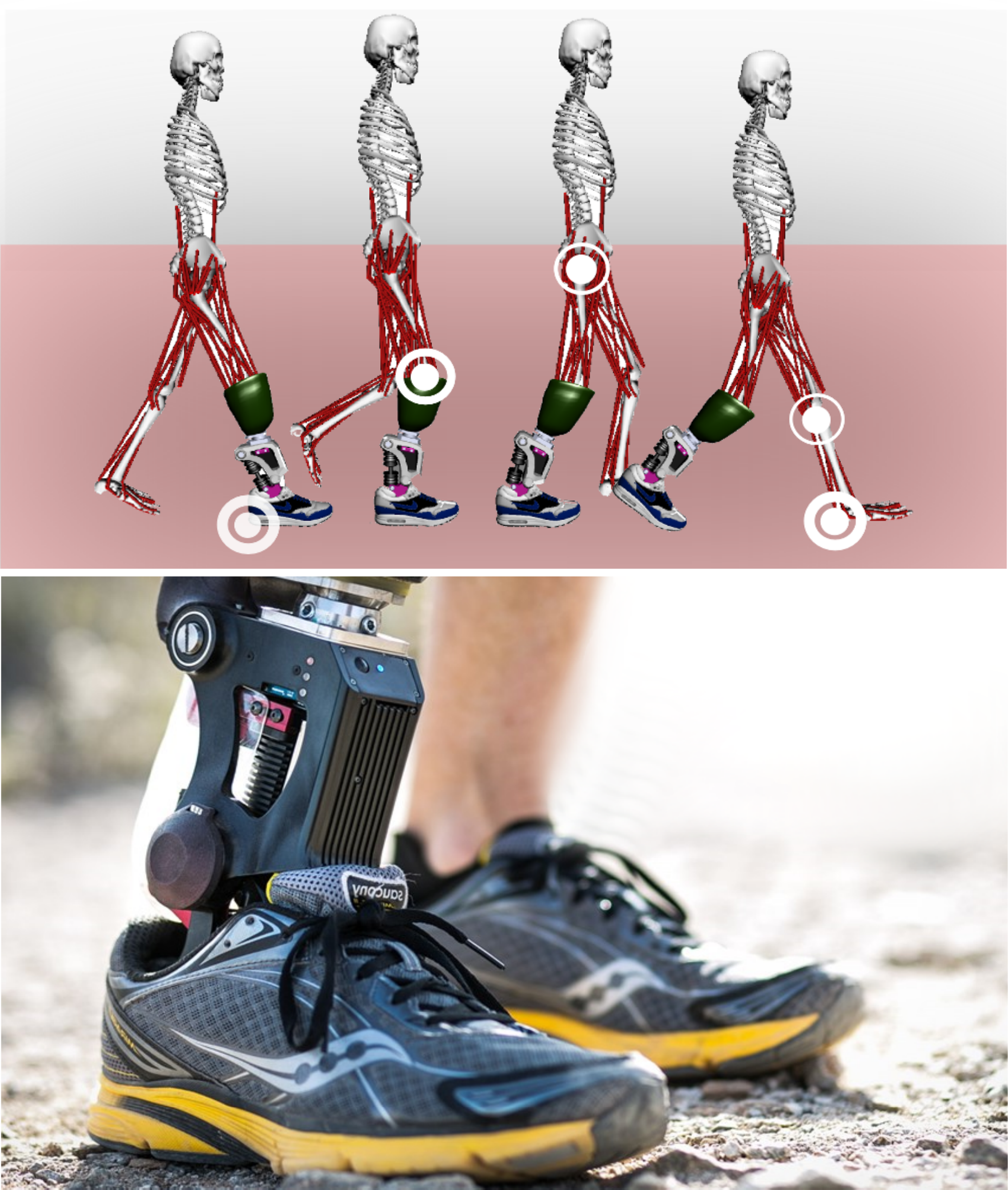}
\caption{Top: Probabilistic modeling predicts upcoming motion and joint stresses. Circles indicate high-impact joints. Bottom: Control signals for our robotic ankle prosthesis are generated in order to minimize joint stress.}
\label{fig:teaser}
\vspace*{-\baselineskip}
\end{wrapfigure}
\emph{Intuitive and safe mobility} is a necessary part of everyday life, and a critical challenge to those with musculoskeletal conditions and lower-extremity amputations~\citep{sinha2011factors, deans2008physical}.
According to the United States Bone and Joint Initiative~\citep{USBJI}, a record number of 126.6 million Americans are affected by a musculoskeletal condition. Out of these, about 2 million are living with a limb loss -- a number that is likely to double by the year 2050~\citep{sheehan2014impact}.
To date, amputees who are longtime prosthesis users are at a high risk for knee osteoarthritis (OA)~\citep{MORGENROTH2012S20} in the \emph{intact} limb -- even when modern microprocessor-controlled prostheses are used. 
By favoring the intact limb, users may develop asymmetric gaits thereby generating higher knee loads in the healthy leg.
Over time, the increased stresses accumulate causing secondary musculoskeletal conditions, further compounding any disability. 

Prostheses, orthoses, exoskeletons, and other assistive robotics devices~\citep{herr2009exoskeletons} have the potential to substantially improve the quality of life of persons with amputations and other musculoskeletal conditions. 
However, to ensure the efficacy and long-term safety of such devices, it is critical to develop control strategies that respond to the ergonomic needs of each specific user. More specifically, forces, torques or moments acting on the human body need to be explicitly taken into account during action generation in order to achieve user-specific, ergonomic, and biomechanically-safe control. 

In this paper, we identify assisted mobility as a problem of symbiotic interaction between a human and assistive robot.
In such a scenario, it is imperative that the assistive device learns models that can anticipate the future evolution of the interaction and adapt to their specific human partner.
To this end, we discuss a unified approach to (a) human state estimation and prediction, (b) inference of unobserved biomechanical variables, and (c) model-predictive control in support of the above ergonomic objectives and constraints.
Building upon existing literature and prior work on probabilistic human motion modelling~\citep{Benamor-RSS-19, geoffrey2020predictive}, we present the application of interaction primitives for learning predictive models of biomechanics and control in a lower-limb prosthesis.
An important feature of this method is the ability to reduce the cooperative system  (human and robot) to a latent space using basis function decomposition, in which temporal features are probabilistically related to one another through their covariance.
We leverage nonlinear Bayesian filtering, to forecast the future evolution of a multi-modal walking or jumping time-series.
At runtime, different types of predictions about the future state of the human user,.e.g., the forces acting on different body parts, can be made. Optimal actions are, in turn, chosen in a model predictive control loop so as to meet a specified objective, i.e., minimize knee forces over a finite time horizon.
To demonstrate prediction and control abilities of our method, we present a variety of simulated and real-world experiments that involve the real-time, ergonomic control of a commercial lower-limb prosthesis.
\begin{figure}[t]
    \centering
    \includegraphics[width=0.8\columnwidth]{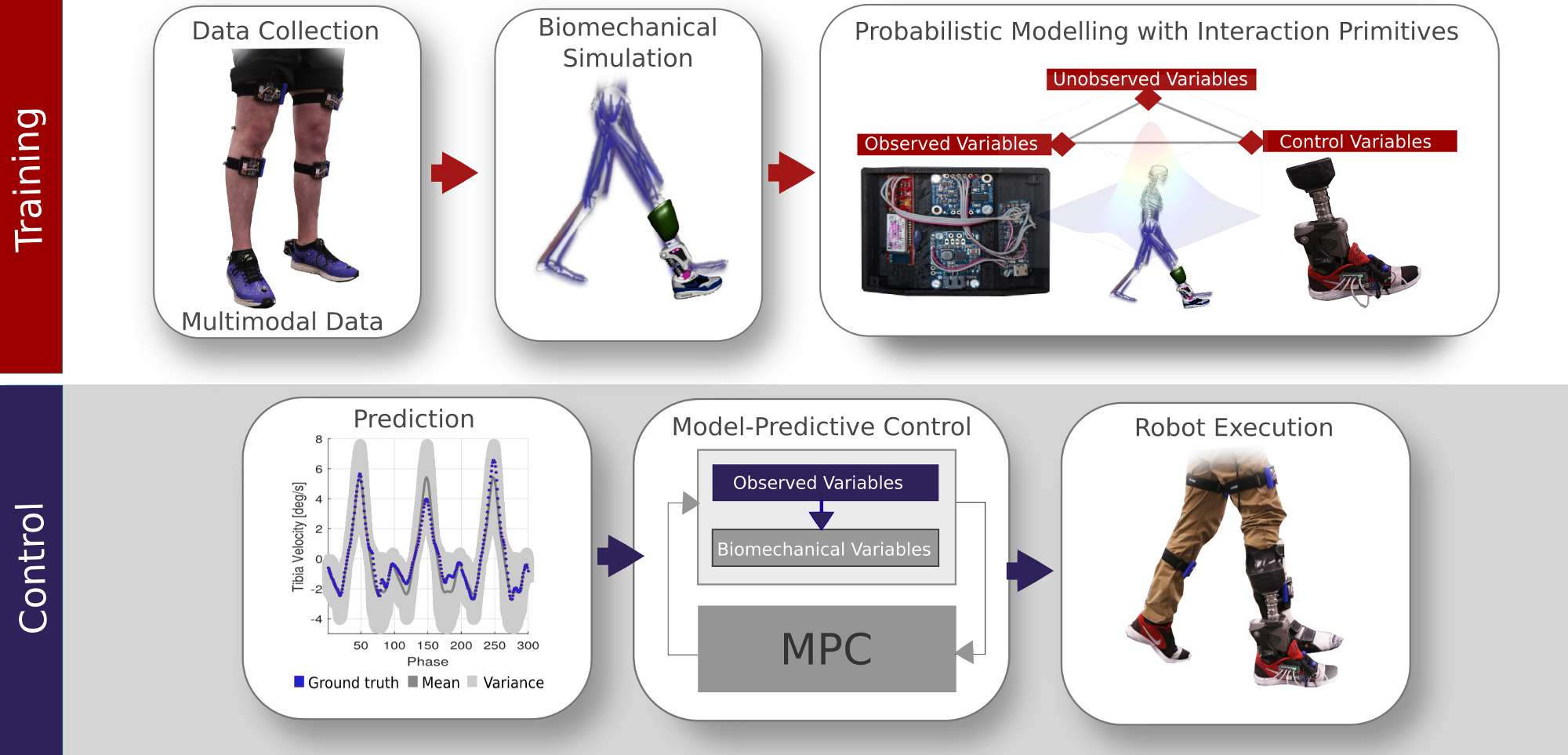}
    \caption{Overview diagram of our approach depicting the training and deployment phases.}
    \label{fig:overview}
    \vspace*{-\baselineskip}
\end{figure}

\section{Related Work}
In recent years, we have seen substantial progress in the domain of wearable robotic devices such as prosthetics, orthoses, and exoskeletons~\citep{herr2009exoskeletons}. However, user-adaptation and ergonomy still remains a major challenge~\citep{tucker2015control}. Machine learning techniques are well positioned to overcome this challenge by enabling new forms of user-adaptation and customization. The work in~\citep{gao2020recurrent} builds upon the recent surge in deep learning approaches and describes end-to-end learning of continuous control with a recurrent neural network. Neural network approaches have a remarkable predictive power and are particularly popular for anticipating (discrete) events and walking conditions such locomotion mode detection~\citep{alharthi2019deep}, human intention inference~\citep{su2019cnn}, or environment recognition~\citep{khademi2019convolutional}. Despite recent developments in certification, monitoring, and verification techniques~\citep{katz2017reluplex, dokhanchi2018evaluating}, there are still major barriers for the application of deep networks in safety-critical systems such as prosthetics. Chief among these barriers is the difficulty in providing safety certifications --  a challenge exacerbated by opportunities for adversarial attacks~\citep{su2019one}.

Alternatively, a number of approaches for incorporating machine learning into prosthetic control are based on probabilistic and Bayesian formulations. For example, the work in~\citep{Thatte-2019-120186} leverages Gaussian processes (GPs) to learn a predictive model for hip angles and hip heights during walking.
Gaussian processes have been particularly popular in locomotion research~\citep{lizotte2007automatic, calandra2016bayesian, atkeson2016, rastgaar2018}, due to their ability to learn from small sample sizes, their ability to produce posterior distributions with uncertainty information, and the relative ease with which they can be integrated into existing (often Kalman-based) filtering and control frameworks. 

Another probabilistic framework termed Interaction Primitives(IPs)~\citep{amor2014interaction, campbell2017bayesian}, combines dynamical movement primitives~\citep{ijspeert2013dynamical} with a Bayesian inference step, to produce an imitation learning approach specialized for modeling the interaction of multiple agents. The IP combination of a.) parameterizing models of two interacting agents via basis function decomposition, followed by b.) jointly modeling the trajectory distributions (utilizing the basis covariance), and c.) recursive Bayesian filtering (Kalman or Particle filtering); is particularly well-suited for predictive modeling of human motion. Previous applications of IPs for human-robot interaction include, for example, collaborative assembly~\citep{ewerton2015}, throwing and catching~\citep{Benamor-RSS-19}, or electromyography based object hand over~\citep{chen2017learning}, show that IPs can infer modeled parameters including robot joint angles given a set of observed partial trajectories. A first attempt at using IPs for robotic prostheses is presented in~\citep{geoffrey2020predictive}, where control signals were generated in a purely reactive fashion by iteratively conditioning on the sensor readings. The approach is particularly powerful when dealing with noisy sensing and performance gracefully degrades when one or more sensors unexpectedly break down; however it does not take into account the future ramifications of a selected action.  Actions of a prosthesis that are locally optimal, may potentially lead to trips, high internal stresses, or other dangerous situations. 

In this paper, we build upon previous work on IP, but make critical extensions to allow for planning and biomechanically-safe, predictive control. Going beyond reactive control~\citep{ewerton2015, geoffrey2020predictive, Benamor-RSS-19}, we show how model-predictive control (MPC)~\citep{mayne2014model,qin2003survey} can be implemented with IPs as the underlying data-driven model. 
The resulting approach has powerful inference capabilities that allow for \emph{predictive biomechanics}: the prediction and control of future, unobserved biomechanical states of the human-robot system from current sensor readings. Furthermore, the approach plans robot control trajectories that minimize biomechanical stress on the human body, and we show that the underlying cost functions specifying biomechanical safety can be modified to realize different ergonomic goals.

\section{MPIP: Model-Predictive Interaction Primitives}
Our approach, illustrated in Fig.~\ref{fig:overview}, consists of two phases and can be seen as an imitation learning process~\citep{schaal1999imitation}. In the training phase, we record training data from the process we are interested in modelling, i.e., human locomotion. This step typically involves recording data from a set of input modalities, e.g., accelerometers, force plates, motion capture etc. The raw data set may not yet include variables that are of interest but are not immediately measurable, such as knee loads, muscle forces and the control signals for the prosthesis. These additional variables are acquired through simulations and incorporated into the data set. In turn, the augmented data set is used for training an Interaction Primitive --  a probabilistic model of the underlying system dynamics~\citep{amor2014interaction, ewerton2015, campbell2017bayesian}. After training, the real-time controller running on the assistive device uses the observed sensor readings as input to predict future sensor values, biomechanical variables, and reactive control signals. Within a model-predictive control loop, predictions are made on how different robot actions will affect the evolution of these values. Via MPC we optimize an action that best satisfies a given cost function, e.g., the total muscle work, the knee forces or other criteria essential to healthy motion.
IPs can effectively be cast as a linear model in which all nonlinearity is solely encoded within a set of basis functions. In the following sections, we will describe each step of our methodology in more detail.

\subsection{Data Collection and Augmentation}\label{sec:augmentation}
The presented method is inherently an imitation learning technique with planning capabilities. As such, it must be trained with example demonstrations of the behavior to be learned.
For data collection, a human user is equipped with a diverse set of sensors. These may include accelerometers, gyros, pressure-sensitive shoes, and other sensors that provide information about the state of the user. The necessary sensor data representing the subject's state is collected from $N$ individual demonstrations of an action, $[ \mb{y}_1, \dots ,\mb{y}_{T_n} ] \in \mathbb{R}^{D_y \times T_n}$, such that $D_y$ is the number of sensor signals and $T_n$ is the number of time steps in the $n^{\text{th}}$ demonstration. In order to cover a variety of settings, demonstrations should ideally cover multiple gait cycles, speeds and ground inclinations. However, these values do not yield a sufficiently detailed picture of the walking process. In particular, non-observable biomechanical variables are missing. Examples are joint forces, muscle effort and mechanical work. These can be calculated via analytical methods for which a number of musculoskeletal simulation libraries are available, e.g., OpenSim~\citep{delp2007opensim}. 
In our specific scenario, we use  pressure-sensing shoes to measure the ground reaction forces at four points and, in turn, estimate joint reaction forces through the Newton-Euler formulation of 3D inverse dynamics as described in~\citep{robertson2013research}. External forces acting on the foot are reduced to a single equivalent force which represents the sum of all effective forces acting on the body. Consequently, we project the one-dimensional foot force from the smart shoe into a 3D force by assuming that the force originates at the center of pressure (COP) in the shoe and traverses a straight line through the divergence point located just above the subject's center mass (CM). Newton-Euler equations are then used to calculate the force and moment at the proximal end of each segment up to the CM. 
 
We follow best practices to ensure sufficiently accurate approximation of dynamics. In particular, we calibrate with accurate physiological measurements from the user (see supplemental material), incorporate sanity checks for expected range of values, and use the best practices in \citet{hicks2015simulation}. Most importantly, we cross-compare the results from OpenSim with results generated from motion capture and force plate technology (Accugait, AMTI, Watertown, MA, USA) to ensure a match between the results ($\leq5\%$ error). While such force plate technology is the de facto standard in force calculation and used for clinical gait analysis, it is static and cannot be built into the shoes. The work in \citet{nejad2020capacity} noticed ``that current generic musculoskeletal modelling techniques are able to reproduce the in vivo conditions (i.e. ground truth) measured during walking" which is one of the main targets of the paper. Please also note, that while not standard today, technology exists to measure such forces in vivo \citep{d2013implantable}. Our approach is agnostic to the type of sensor and can also be used with these in vivo sensors instead of simulation data.

The final step in data augmentation is to provide robot control signals. These signals can either be recorded from within the demonstration, i.e., setting the robot's target angle to the human ankle ankle; or can be developed through a modeling process such as a complex impedance control system. The objective of this step is to provide an approximate control signal which is later refined through model-predictive control. Using the above methodologies, trajectories of latent variables (biomechanical variables and control signals) $[\mb{m}_1, \dots ,\mb{m}_{T_n}] \in \mathbb{R}^{D_m \times T_n}$ are incorporated into the dataset where $D_m$ is the number of latent variables. The full data matrix for a demonstration then consists of $\mb{Y}_{1:t} = [\mb{y}_t, \mb{m}_t] \in \mathbb{R}^{D}$, where $D = D_y + D_m$.

\subsection{Preliminaries: Latent Space Formulation and Inference}
\label{sec:learning interaction}
\textbf{Latent Space Formulation:} We approach imitation learning from a probabilistic perspective, with the high-level goal of estimating a distribution over future observed and unobserved state variables $\hat{\mb{Y}}_{t+1:T}$ given previously observed states $\mb{y}_{1:t}$ and a prior set of $N$ demonstrations $\mb{Y}^1_{1:T_1}, \dots, \mb{Y}^N_{1:T_N}$:
\begin{align}
    p(\hat{\mb{Y}}_{t+1:T} | \mb{y}_{1:t}, \mb{Y}^1_{1:T_1}, \dots, \mb{Y}^N_{1:T_N}).
    \label{eq:ip_general}
\end{align}
This formulation, however, would require a nonlinear transition function for observed and unobserved variables, as well as the definition of a generative distribution over them, which is a non-trivial challenge.
An existing strategy~\citep{campbell2017bayesian} is to transform the state representation into a time-invariant latent space via basis function decomposition.
In this method, basis functions capture the evolution of the state variables over time while the relationship between basis functions is modeled with a joint probability distribution.
We define the basis decomposition as a linear combination of weighted basis functions which approximates each state dimension $d$ as $\mb{Y}^d_{t} = \mb{\Phi}_{\phi(t)}^{\intercal} \mb{w}^d + \epsilon_y$ where $\mb{\Phi}_{\phi(t)} \in \mathbb{R}^{B^d}$ is a vector of $B^d$ basis functions, $\mb{w}^d \in \mathbb{R}^{B^d}$ is a vector of corresponding weights and acts as our latent state variable, $\phi(t) \in \mathbb{R}$ such that $0 \leq \phi(t) \leq 1$ is a relative measure of time known as \emph{phase} (hereafter denoted as $\phi$), and $\epsilon_y$ is the approximation error.
The choice of basis function depends on the data and in this work we opt to utilize a Gaussian function, however both von Mises as well as polynomial basis function are viable given the domain.
The full latent state representation is given by the concatenation of all latent state vectors, $\mb{w} = [\mb{w}^{0\intercal}, \dots, \mb{w}^{D\intercal}] \in \mathbb{R}^{B}$ where $B = \sum_d^D B^d$, together with the phase $\phi$ and phase velocity $\dot{\phi}$ such that $\mb{s} = [\phi, \dot{\phi}, \mb{w}]$~\citep{campbell2017bayesian}.
Since $\mb{w}$ is time-invariant, estimating this quantity is equivalent to estimating all past, present, and future state values of $\mb{Y}$; thus we can reformulate Eq.~\ref{eq:ip_general} as $p(\mb{s}_t | \mb{y}_{1:t}, \mb{s}_{0})$.
After application of Bayes' rule, this yields the following posterior distribution:
\begin{equation}
\label{eq:ip_general2}
p(\mb{s}_t | \mb{y}_{1:t}, \mb{s}_{0}) \propto p(\mb{y}_{t} | \mb{s}_t) p(\mb{s}_t | \mb{y}_{1:t-1}, \mb{s}_{0}). 
\end{equation}
\textbf{Inference:} We employ an approximate inference scheme found in ensemble Bayesian Interaction Primitives~\citep{campbell2019learning} solve the posterior distribution in Eq.~\ref{eq:ip_general2}. Assuming that a) the Markov property holds and b) errors in the state estimate are normally distributed (that is, we ignore higher-order statistical moments), the posterior defined by Eq.~\ref{eq:ip_general2} can be solved with a recursive two-step Bayesian filter in which the distributions are approximated with an ensemble of state members. We define an ensemble $\mb{X}$ consisting of $E$ members such that $\mb{X} = [\mb{x}^1, \dots, \mb{x}^E]$. The initial ensemble, $\mb{X}_0$, is sampled directly from the prior: $\mb{x}_0 \sim p(\mb{s}_0)$ for all $\mb{x}_0 \in \mb{X}_0$. In practice, since we do not know $p(\mb{s}_0)$ it is common to set $E = N$ and utilize the demonstration latent states directly as samples.
The first step approximates $p(\mb{s}_t | \mb{y}_{1:t-1}, \mb{s}_{0})$ by propagating each ensemble member forward in time with
\begin{align}
\mb{x}^j_{t|t-1} &=
g(\mb{x}^j_{t-1|t-1})
+
\epsilon_x, \quad 1 \leq j \leq E,
\label{eq:state_prediction}
\end{align}
where $g(\cdot)$ is a constant-velocity transition operator and $\epsilon_x$ is a noise perturbation. The second step updates the members with the observation, leveraging the nonlinear observation operator $h(\cdot)$:
\begin{align}
\mb{H}_t\mb{X}_{t|t-1} &= \left[h(\mb{x}^1_{t|t-1}), \dots, h(\mb{x}^E_{t|t-1})\right]^\intercal,\\
\mb{H}_t\mb{A}_t &= \mb{H}_t\mb{X}_{t|t-1}
- \left[ \frac{1}{E} \sum_{j=1}^{E}h(\mb{x}^j_{t|t-1}), \dots, \frac{1}{E} \sum_{j=1}^{E}h(\mb{x}^j_{t|t-1}) \right]. \nonumber
\end{align}
As a result, no linearization error is introduced as would be the case in analytical formulations such as the EKF. The innovation covariance can now be found with
\begin{align}
\mb{S}_t &= \frac{1}{E - 1} (\mb{H}_t\mb{A}_t) (\mb{H}_t\mb{A}_t)^\intercal + \mb{R},    
\end{align}
where $\mb{R}$ is the observation noise matrix.
This is then used to compute the Kalman gain as
\begin{align}
\mb{A}_t &= \mb{X}_{t|t-1} - \frac{1}{E} \sum_{j=1}^{E}\mb{x}^j_{t|t-1}, \qquad
\mb{K}_t = \frac{1}{E - 1} \mb{A}_t (\mb{H}_t\mb{A}_t)^\intercal \mb{S}^{-1}_t.
\end{align}
Finally, the ensemble is updated to incorporate the new measurement perturbed by stochastic noise:
\begin{align}
\mb{\tilde{y}}_t &= \left[ \mb{y}_t + \epsilon^1_y, \dots, \mb{y}_t + \epsilon^E_y \right], \qquad
\mb{X}_{t|t} = \mb{X}_{t|t-1} + \mb{K} (\mb{\tilde{y}}_{t} - \mb{H}_t\mb{X}_{t|t-1}).
\label{eq:measurement_update}
\end{align}

It is important to note that during inference observations can be partial.
In our case, we leverage this ability to generate a posterior over biomechanical variables despite the fact that they cannot be immediately observed.
This follows from the data augmentation strategy employed in training.

\subsection{Model-Predictive Control with Interaction Primitives}
\label{sec:mpc}
Given the trained probabilistic model which can predict the temporal evolution of the system, we can construct an efficient model predictive control (MPC) framework. MPC~\citep{qin2003survey} computes an open-loop optimal control strategy over a given time horizon. The loop is, then, closed by instantiating the first control value and periodically recomputing the control strategy from an updated state. As opposed to the traditional state-based sampling method, which predicts the time horizon trajectory by iteratively calling the state transition function, our probabilistic model inherently contains the future state prediction for observed as well as latent signals.
We start with a cost function formulation to find the optimal control strategy $\mb{u}^*$:
\begin{equation}\label{equ:MPC_general}
      J(\mb{u}^*) = \int_{t}^{H_x}||\mb{x}^u||^2dt + \int_{t}^{H_u}||\Delta \mb{u}||^2dt~,
\end{equation}
where the cost function $J$ takes the integral of a variable-of-interest time vector $\mb{x}^u$ with respect to the control output $\mb{u}$, as well as the change in the control output vector $\Delta \mb{u}^* = \mb{u} - \hat{\mb{u}}$, shown as the difference between the control signal to be optimized $\mathbf{u}$ and the estimated control signal from the IP process $\hat{\mathbf{u}}$.
Moreover $H_x$ and $H_u$ indicate the signal and control horizons respectively. To make use of the probabilistic model, we first transform the cost function into the phase domain by integrating from the current phase to the time horizons in phase $H_{x_\phi}$ and $H_{u_\phi}$.
The benefit of utilizing a basis function model is that, once in the phase domain, we can easily extract the weight vector $\mb{w}$ out of the integrals, since we represent the full trajectory $\hat{\mb{Y}}^d_{1:P} = [\bs{\Phi}^{\intercal} \hat{\mb{w}}_t^d]$ as the inner product of basis vector and the weight vector, resulting in:
\begin{equation}
    J(\mb{u}^*) = \int_{\phi}^{H_{x_\phi}}||\bs{\Phi}^{\intercal}||d\phi ~ ||\mb{w}_{x^u}||^2 + \int_{\phi}^{H_{u_\phi}}||\bs{\Phi}^{\intercal}||d\phi ||\Delta\mb{w}_u||^2~,
\end{equation}
The weight vector for the variable-of-interest (e.g. knee forces, muscle forces) is simply the possible control strategy times the learned prior model, namely the covariance between the weight vectors of the control and variable-of-interest $\mb{w}_{x^u} = \text{Cov}(\mb{w}_{x}, \mb{w}_{x}) \cdot \mb{w}_{u^*} $

For computational speed-ups, the analytical integral $\bs{\Psi}_{\phi:H}$ can be pre-calculated to further reduce the number of computations. In the case of the Gaussian basis function, integration reduces to:
\begin{align}
    \bs{\Phi}(\phi_{0:1}) = \frac{1}{\sigma\sqrt{2\pi}}e^{-(\frac{\phi-\mu}{2\sigma})^2} \\
    \bs{\Psi}_{\phi:H} = \int_{\phi}^{H}||\bs{\Phi}(\phi_{0:1})||^2 = \sqrt{\frac{2}{\pi}}e^{\frac{\mu-\phi}{2\sigma}}
\end{align}
resulting in a simple and easy to pre-compute closed-form integral that does not require stochastic sampling. Building on this result, the formulation of the optimization problem for generating controls thus becomes:
\begin{equation}
    \mb{u}^* = \text{argmin}( \bs{\Psi}_{\phi:H_{x_\phi}}^\intercal  ||\mb{w}_{x^u}||^2 + \bs{\Psi}_{\phi:H_{u_\phi}}^\intercal ||\Delta\mb{w}_u||^2).
\end{equation}
Solutions to the above optimization problem can be found using modern constrained optimization solvers such as Sequential Least Squares Programming (SLSQP)~\citep{kraft1988software}. We impose constraints on the controller by using the the min and max values of each basis function from the training demonstrations.
Alternatively, one can also use a multiobjective MPC solver to compute a Pareto-optimal solution, see \citet{bemporad2009multiobjective}.

\section{Experiments}
In this section, we present a set of experiments that highlight the capabilities of the introduced method. We our method to the target domain of assisted mobility in both simulation and real-world.

\subsection{Experiment 1: Simulation}
In the first experiment, we apply our MPIP method to the problem of prosthesis control for assisted mobility. To this end, we build upon the NeurIPS ``AI for Prosthetics" competition in 2018. The competition featured a passive prosthesis (not controlled) and focused on controlling the virtual human through muscle activations. The problem that is at the center of our paper, however, is the challenge of actively controlling a prosthesis so as to best adapt to a human user (not controlled). Accordingly, we modified the simulation environment by adding a controllable degree-of-freedom for powered actuation at the ankle of the prosthesis. The NeurIPS competition simulator uses OpenSim in order to provide a physiologically-accurate musculoskeletal model, along with a physics-based environment, and various data sets of gait kinematics. For training, we used the winning submission of 2018 and incorporated low-frequency random control signals to the prosthesis. More specifically, we adopt a quasi-active control scheme for the prosthesis -- a popular approach for modern powered-ankle prosthetic devices. In this control strategy, the offset of the spring attached to the prosthesis is set to mimick the quasi-static stiffness curve of an intact ankle during walking~\citep{au2007powered}. 
\begin{table}[!b]
\vspace*{-\baselineskip}
\caption{Results of walking experiment in simulation.}
\resizebox{\textwidth}{!}{%
\begin{tabular}{lccccc}
\hline
\hline
\multicolumn{6}{c}{Simulation Experiment 60 Stride Average} \\
\cline{1-6}\cline{1-6}
\multicolumn{1}{c}{} & Medial Knee Impulse &  Vertical Knee Impulse & Muscle Impulse & Prosthetic Impulse & Dynamic Stability\\
\multicolumn{1}{c}{Control Methods} & (Nm$\cdot$ s) & (N$\cdot$ s) & (N$\cdot$ s) & (N$\cdot$ s) & Lyapunov Exponent \\
\hline
\multicolumn{1}{l}{Passive Prosthesis} & 44.6 $\pm$ 16.3 & 1114 $\pm$ 209 & 1880 $\pm$ 337 & 25.1 $\pm$ 4.01 & 1.78  \\
\multicolumn{1}{l}{Interaction Primitives~\citep{geoffrey2020predictive}}     & 44.9 $\pm$ 23.9 & 1160 $\pm$ 275 & 1880 $\pm$ 445 & 22.3 $\pm$ 3.31 & -0.617 \\
\rowcolor{Gray} \multicolumn{1}{l}{MPIP (Knee Force Reduction)} & \tb{29.4} $\pm$ 12.5 & \tb{913} $\pm$ 245 & 1900 $\pm$ 238 & 21.3 $\pm$ 4.82 & \tb{-0.226} \\
\multicolumn{1}{l}{MPIP (Muscle Force Reduction)} & 47.6 $\pm$ 19.2 & 1200 $\pm$ 186 & \tb{1680} $\pm$ 469 & 23.9 $\pm$ 2.45 & 3.31 \\
\hline
\hline
\end{tabular}\label{ref:tableSim}}
\vspace{-0.5cm}
\end{table}
We used the above controller, along with biomechanical information including: a.) Vertical Knee Reaction Force - sum on all external forces from simulation including body kinetics and muscle forces in the vertical direction, and
b.) Sum of Muscle Forces - summation of the right leg muscle groups available in the simulation environment including: hamstring, biceps femoris, vasti muscles, rectus femoris, iliopsoas, and gluteus maximus; gathered from OpenSim in order to train an IP of the underlying walking behavior. After training, we performed walking experiments in which the virtual agent repeatedly performs 60 strides with 3 different modes of our controller. In the first mode, we only use the inference mechanisms of IPs to generate reactive control values. This can be seen as a behavioral cloning approach and corresponds to the method introduced in~\citep{geoffrey2020predictive}. In addition we evaluated our MPIP variant with a cost function that minmizes a.) the vertical knee reaction force and b.) the sum of muscle forces. The results of this experiment can be seen in Tab.~\ref{ref:tableSim}. The passive prosthesis corresponds to the winning NeurIPs submission with a powered ankle described above. We can observe that the reactive IP approach behaves similar to the passive prosthesis in regards to the internal stresses (e.g. knee impulse). These results are to be expected since the IP effectively clones the controller behavior. Interestingly, however, an improvement in the Local Dynamic Stability (LDS) can be observed. Empirically, we noticed that this is likely due to the fact that the typical IP approach executes the \emph{mean} control signal of the generated posterior. As a result, some stability ``buffer'' is induced by the variance around the mean prediction.
Tab.~\ref{ref:tableSim} also shows the results for our MPIP method, when optimizing knee force. A significant reduction in the medial and vertical knee impulse can be observed. In addition, a significant improvement to the LDS Lyapunov exponent can be observed, which corresponds to higher stability and reduced risk of falls. The LDS metric has been shown to be a reliable measure for assessing dynamic stability of locomotion in pathological and healthy populations \citep{ekizos2018}. Our experiment also shows a significant reduction in the muscle impulse when optimizing for muscle force using MPC. In this case, however, this impulse reduction did not also translate to an improvement in the LDS. Rather, the optimization of muscle force lead to a significant reduction in stability. It can be argued that higher muscle forces are required for resilient responses to external perturbations.
\begin{figure}[!t]
    \centering
    \includegraphics[width=1\textwidth]{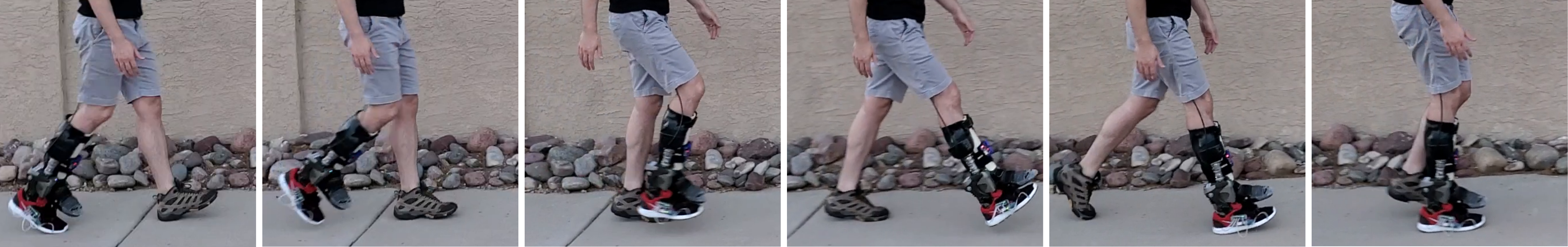}
    \caption{Assisted mobility with model-predictive interaction primitives (MPIP) running in real-time on a SpringActive Ruggedized Odyssey prosthesis. All control is handled by our method.}
    \label{fig:realWalkSidewalk}
    \vspace*{-\baselineskip}
\end{figure}

\subsection{Experiment 2: Real-World Prosthetic Experiments}
Next, we performed real-world experiments using the ``Ruggedized Odyssey" an ankle prosthesis by SpringActive~\footnote{\url{https://springactive.com/}} which incorporates a high power motor and parallel-elastic mechanism. To this end, the different control strategies discussed in this paper were implemented on the embedded controller of the prosthesis. An able-bodied human subject was asked to perform 20 jumps, 10 examples for soft landing and 10 examples for hard landing with more impact on the ground, for training (a bypass connected to the prosthesis was used to attach it to the participant's leg). After training, the same participant was asked to repeat 10 trials with different control strategies running on the prosthesis. Tab.~\ref{ref:realWorld} presents the results of this experiment. Besides the reactive IP inference, we also evaluated our MPIP approach with a cost function for a.) reducing knee forces (minimization), b.) increasing knee forces (maximization), and c.) increasing symmetry of the ankle angles between both legs. 
\begin{wrapfigure}{r}{0.35\textwidth}
      \centering
    \vspace*{-.5\baselineskip}
    \includegraphics[width=.35\textwidth]{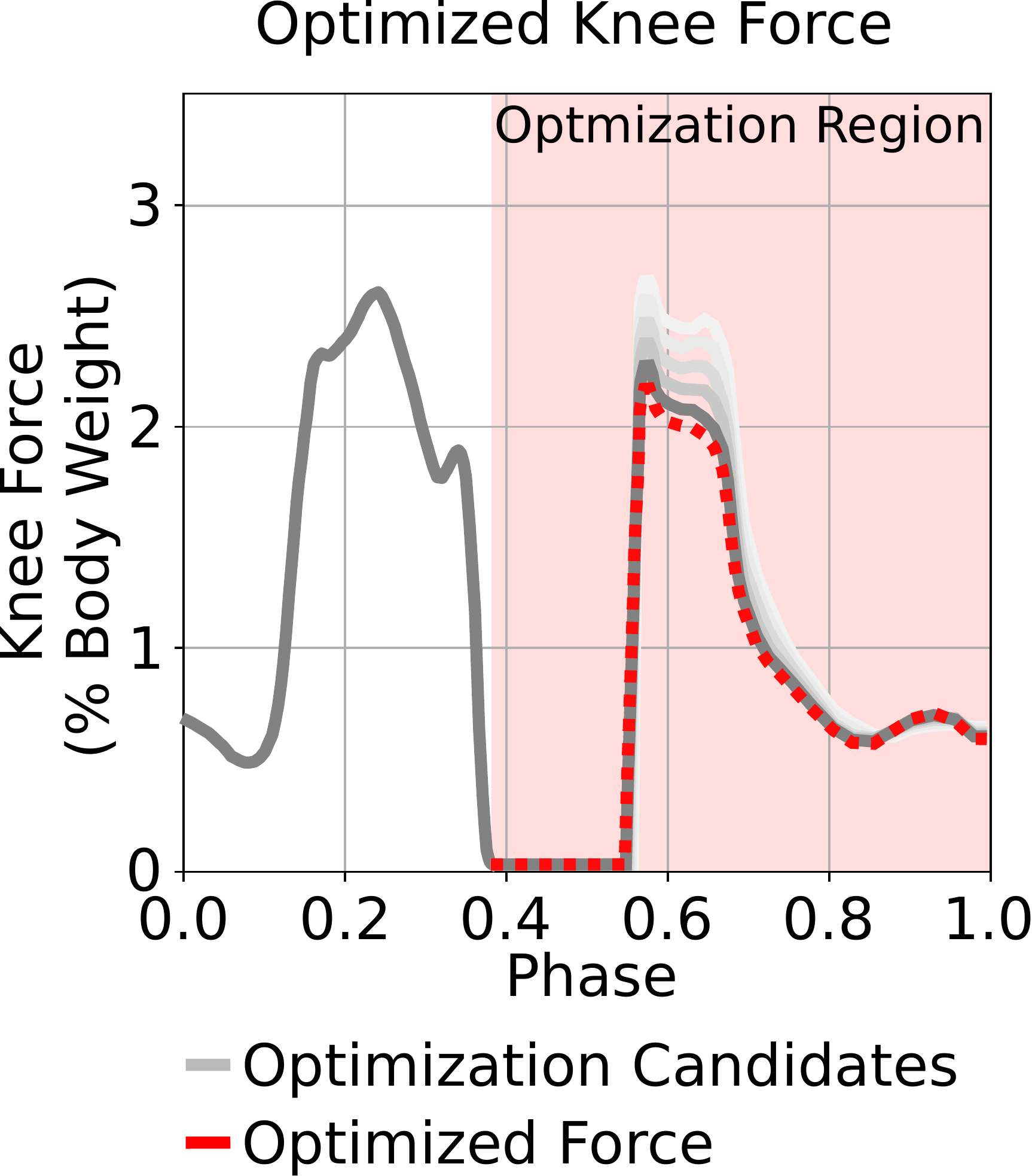}
    \caption{MPC resulting in reduced knee forces.}
    \label{fig:optimization}
    \vspace*{-.5\baselineskip}
    \vspace*{-\baselineskip}
\end{wrapfigure}
We can observe, that the lowest knee force is achieved when using the MPIP Reduction strategy. Fig.~\ref{fig:optimization} depicts the MPC optimization process and visualizes a number of candidate solutions and the resulting minimal force profile. 
Interestingly, the peak toe forces increase, indicating a shift in force distribution towards the metatarsal area. The MPIP Symmetry strategy also implicitly leads to a reduction in knee forces -- an interesting observation that confirms the importance of maintaining symmetric gaits in order to reduce the risks of osteoarthritis. 

\begin{figure}[!h]
    \centering
    \includegraphics[width=\textwidth]{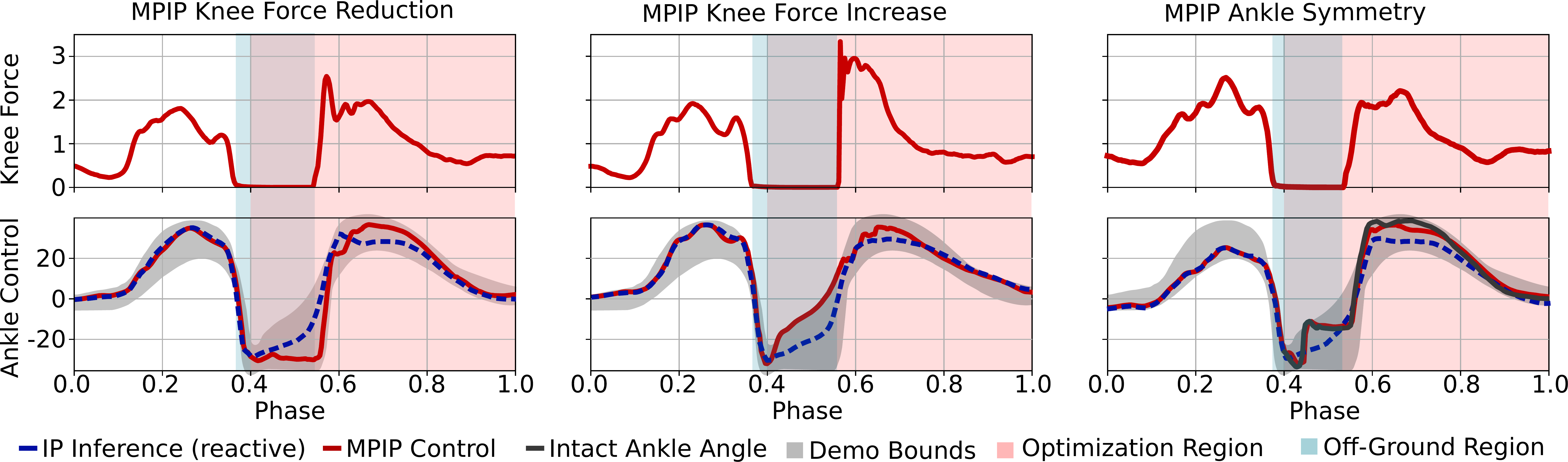}
    \caption{Different control strategies based on changing cost functions. Control signals are chosen to minimize knee loads, maximize knee loads or generate ankle angles for both legs that are symmetric.}
    \label{fig:jumping}
    \vspace*{-\baselineskip}
\end{figure}

An in-depth visualization of the effects of the different strategies can be seen in Fig.~\ref{fig:jumping}. The top row depicts the knee forces for the different strategies. MPIP Increase leads to a substantial increase in the knee force. In contrast to that both MPIP Symmetry and MPIP Reduction maintain a low force profile. In the bottom row we see effects of the strategies on the ankle angle. Substantial differences can be observed between ankle angle trajectories for the knee force reduction and the knee force increase. In the case of symmetry matching (bottom right) we can observe that the generated control trajectory closely matches the ankle angles of the intact leg. While the resulting trajectory substantially differs from the control trajectory for the knee force reduction strategy in flight, both of these options produce a low knee force during landing, i.e., the force profiles in the top row resemble each other, since the subject is able to control both ankles effectively.
Finally, to demonstrate the real-time capabilities of our method, we tested the approach outside of laboratory conditions. In particular, we asked the participant to walk with the prosthesis in a typical outdoor environment, i.e., a street side walk. The MPIP control loop runs at a rate of approx.~\SI{15}{\hertz} and generates smooth walking trajectories for the lower-limb prosthesis. 
Fig.~\ref{fig:realWalkSidewalk} shows an image sequence of the resulting walking gait. 
Hence, the presented method is readily applicable on commercial devices and does not require any changes to the hardware or the computational resources.       
\begin{table}[t!]
\caption{Results of real-world jumping experiment with a SpringActive prosthesis.}
\label{ref:realWorld}
\resizebox{\textwidth}{!}{%
\begin{tabular}{lccccc}
\hline
\hline
\multicolumn{6}{c}{Prosthesis Experiment 10 Trials} \\
\cline{1-6}\cline{1-6}
\multicolumn{1}{c}{} & Peak Heel Force & Peak Toe Force & Peak Knee Force & Total Knee Impulse & Impact Ankle Angle \\ 
\multicolumn{1}{c}{Control Methods} & (Body Weight) & (Body Weight) & (Body Weight) & (Body Weight$\cdot$ s) & (Deg) \\
\hline 
\multicolumn{1}{l}{AB Soft Landing Training} & 0.211 $\pm$ 0.041 & 0.374 $\pm$ 0.093 & 2.36 $\pm$ 0.355 & 118 $\pm$ 16.6 & -22.6 $\pm$ 3.96 \\
\multicolumn{1}{l}{AB Hard Landing Training} & 0.660 $\pm$ 0.045 & 0.406 $\pm$ 0.086 & 3.90 $\pm$ 1.25 & 158 $\pm$ 27.1 & 7.10 $\pm$ 5.22 \\
\hline
\multicolumn{1}{l}{Interaction Primitives}   & 0.333 $\pm$ 0.087 & 0.342 $\pm$ 0.098 & 2.68 $\pm$ 0.687 & 130 $\pm$ 19.6 & -11.6 $\pm$ 4.81 \\
\rowcolor{Gray} \multicolumn{1}{l}{MPIP Reduction} & 0.191 $\pm$ 0.049 & 0.384 $\pm$ 0.095 & \tb{2.26} $\pm$ 0.387 & \tb{109} $\pm$ 13.4 & \tb{-28.6} $\pm$ 2.00 \\
\multicolumn{1}{l}{MPIP Increase} & 0.650 $\pm$ 0.048 & 0.321 $\pm$ 0.096 & 3.49 $\pm$ 0.989 & 146 $\pm$ 22.7 & 9.03 $\pm$ 0.942 \\
\multicolumn{1}{l}{MPIP Symmetry} & 0.511 $\pm$ 0.059 & 3.98 $\pm$ 0.090 & 2.41 $\pm$ 0.764 & 122 $\pm$ 26.0 & -14.8 $\pm$ 8.67 \\
\hline
\hline
\end{tabular}
}
\end{table}

\section{Conclusion}
We present a method for generating ergonomic, biomechanically safe assistive motion in tasks that require physical human-robot collaboration. An important novelty of this approach is the ability to perform predictive biomechanics -- predicting future biomechanical states of the human user given current observations and intended robot actions. In turn, this feature is used within an efficient model-predictive control loop to generate control signals for a commercially available prosthetic to meet user-specific ergonomic requirements, i.e., minimize internal knee loads.

\noindent
\textbf{Limitations and Future Work:} The accuracy of our prediction depends on the quality of the data set. We aim to investigate active learning techniques that improve the model by collecting data after deployment. The learned models in this paper focused on a single behavior and may not perform well in transitions, e.g., walking, running, staircases. We will investigate Bayesian hierarchical methods in order to cover a wide range of locomotive skills within a single model. To guarantee safety at all times, we would like to investigate learning and incorporating barrier functions~\citep{ames2019control} into the MPC process. We recognize that the lower dimensional relationships between observable, control, and latent variables is not a catch-all solution, and individuals with amputations may not fit the general learned representation. However, this can be partially remedied by re-training the approach with data from the specific user, by recording data from people with amputations, or by collecting data from high functioning amputees. We aim to further investigate this aspect going forward. 

\section{Acknowledgments}
This  work  was  supported  by  the  National  Science  Foundation under the career award grant FP00012258, as well as by the Global KAITEKI Center.
\newpage
\bibliographystyle{IEEEtran}
\bibliography{references}

\end{document}